\documentclass[11pt]{article}
\usepackage{coling2020}
\usepackage{times}
\usepackage{url}
\usepackage{latexsym}
\usepackage{graphicx}
\usepackage{capt-of}
\usepackage{colortbl}
\usepackage{arydshln}
\usepackage{multirow}
\usepackage{multicol}
\colingfinalcopy 

\title{Exploring and Exploiting Multi-Granularity Representations for \\ Machine Reading Comprehension}

\author{
Nuo Chen \\
 ADSPLAB, School of ECE \\
 Peking University, Shenzhen, China \\
\\\And
  Chenyu You \\
  Department of Electrical Engineering, \\
  Yale University, CT, USA \\
  
  }

\date{}

\begin{document}
\maketitle
\begin{abstract}
Recently, the attention-enhanced multi-layer encoder, such as Transformer, has been extensively studied in Machine Reading Comprehension (MRC). To predict the answer, it is common practice to employ a predictor to draw information only from the ﬁnal encoder layer which generates the \textit{coarse-grained} representations of the source sequences, i.e., passage and question.
The analysis shows that  the representation of source sequence becomes more \textit{coarse-grained} from \textit{fine-grained} as the encoding layer increases.
It is generally believed that with the growing number of layers in deep neural networks, the encoding process will gather  relevant information for each location increasingly, resulting in more \textit{coarse-grained} representations, which adds the likelihood of similarity to other locations (referring to homogeneity). Such phenomenon will mislead the model to make wrong judgement and degrade the performance.
In this paper, we argue that it would be better if the predictor could exploit representations of different granularity from the encoder, providing different views of the source sequences, such that the expressive power of the model could be fully utilized.
To this end, we propose a novel approach called Adaptive Bidirectional Attention-Capsule Network (ABA-Net), which adaptively exploits the source representations of different levels to the predictor.
Furthermore, due to the better representations are at the core for boosting MRC performance, the capsule network and self-attention module are carefully designed as the building blocks of our encoders, which provides the capability to explore the local and global representations, respectively.
Experimental results on three benchmark datasets, i.e., SQuAD 1.0, SQuAD 2.0 and COQA, demonstrate the effectiveness of our approach. In particular, we set the new state-of-the-art performance on the SQuAD 1.0 dataset.

\end{abstract}

\section{Introduction}
Machine reading comprehension (MRC) \cite{DBLP:journals/corr/abs-2106-02182,you2021self,you2020contextualized} is a long-standing task that aims to teach machine how to read and comprehend a given source sequence, i.e., passage/paragraph, then answer its corresponding given questions automatically. It has large amounts of real application scenarios such as question answering and dialog systems.
In recent years, the deep neural networks, including the recurrent neural networks (RNNs) and convolutional neural networks (CNNs) \cite{DBLP:journals/corr/abs-1812-03593,Zhang2019MC2MC} have been introduced to extract representations of source sequence efficiently and make great progress in MRC.
However, for RNNs-based models, the undeniable deficiency is that due to the sequential dependency, its learning ability of long sentences is insufficient and it is unsuitable for parallel training \cite{DBLP:conf/nips/VaswaniSPUJGKP17}; For CNNs-based models, in which the pooling operation incorrectly discards spatial information and does not consider the hierarchical relationship between extracted features \cite{liu2019neural}.


Recently, the attention-enhanced multi-layer encoder, e.g., Transformer \cite{DBLP:conf/nips/VaswaniSPUJGKP17}, ALBERT \cite{lan2019albert}, RoBERTa \cite{DBLP:journals/corr/abs-1907-11692} and XLNet \cite{yang2019xlnet}, which is based solely on attention mechanisms \cite{Bahdanau2015seq2seq} and eliminates recurrence entirely, has been proposed and has established the state-of-the-art in multiple challenging MRC datasets \cite{DBLP:conf/emnlp/RajpurkarZLL16,joshi2017triviaqa,reddy-etal-2019-coqa}. 
\begin{figure}[t]
\centering
\includegraphics[width=1\linewidth]{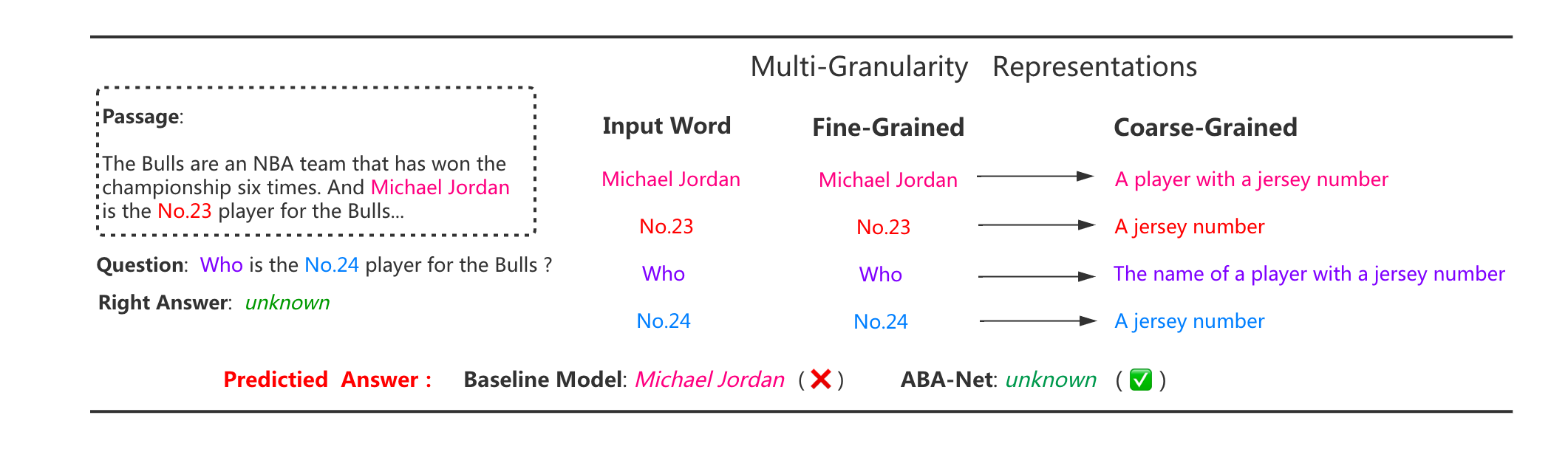}
\caption{An visualization example of a Question-Answer pair for a passage.
  }
  \label{fig:example}
\end{figure}
However, under the multi-layer deep learning setting, the representations of source sequence will become more \textit{coarse-grained} from \textit{fine-grained} with more encoder layers.
Following in \cite{DBLP:conf/iclr/HuangZSC18}, Figure~\ref{fig:example} illustrates that as the multi-layer encoder processes the source sequences, each input word will gradually gather more and more related information as more layers are used, resulting in more abstract representations, i.e., from \textit{fine-grained} to \textit{coarse-grained} representations that are more likely to be similar with other positions (Homogeneous Phenomenon).
For those representations output by different encoder layers, the common practice for the current answer predictor is to draw information (\textit{coarse-grained} representations) only from the final encoder layer. However, it should be intuitive that \textit{coarse-grained} representations are good at expressing the overall meaning of the source sequence, but are less precise in the finer details. If we also exploit \textit{fine-grained} representations for the answer predictor, it will help the predictor find source information more precisely and give answers that are more accurate.
As we observe in Figure~\ref{fig:example}, due to the baseline model only focus on the \textit{coarse-grained} representations, it gives an incorrect answer.
In contrast, our method exploits detailed and accurate information, i.e., the \textit{fine-grained} representations of \textit{NO.23} and \textit{NO.24}, resulting in helping the model focus on the correct source information and predict correct answer.



\begin{figure}[t]
\centering
\includegraphics[scale=0.14]{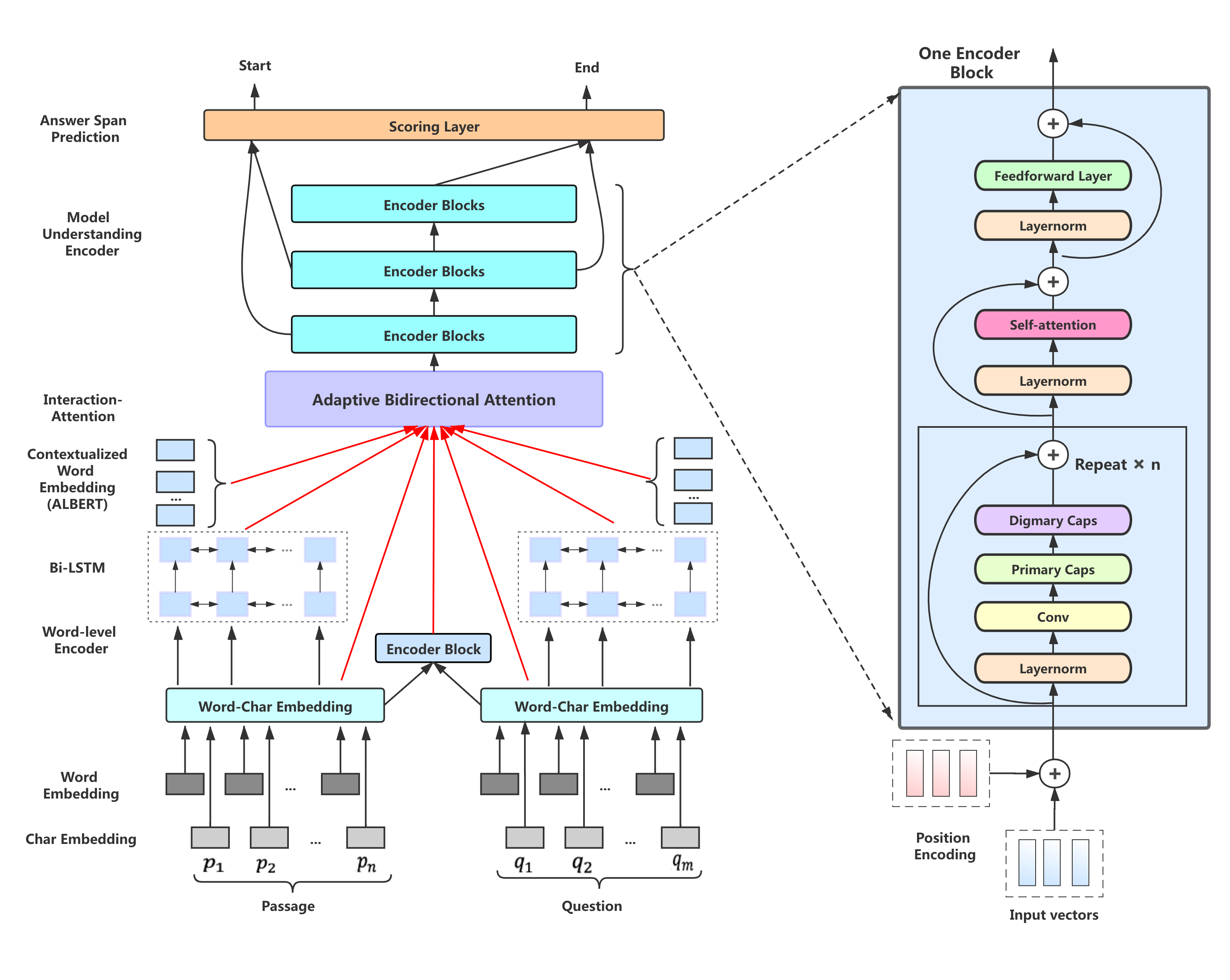}\\
\caption{An overview of the ABA-Net architecture (left). ABA-Net employs Adaptive Bidirectional Attention to reasonably exploit multi-granularity representations (red lines), and specific implementation details can be seen in the Section \ref{sec:interaction-attention}. We use the same Encoder Block (right) throughout the model, only varying the number of Conv-Pri-Dig and layernorm layers for each block. Notice that, we use depthwise separable convolutions \cite{DBLP:conf/cvpr/Chollet17} instead of traditional ones.  Following  \cite{DBLP:journals/corr/abs-1804-09541}, we use a residual connection between each layer
of the Encoder Block and share weights of the passage and question encoder in the embedding  layer, as well as the three 
modeling encodes .
  }
  \label{fig:model}
 \end{figure}

In this paper, we propose a novel framework called   Adaptive Bidirectional Attention-Capsule Network (ABA-Net) to dynamically provide multi-granularity source representations for better predicting the correct answer.
As shown in Figure~\ref{fig:model}, the proposed approach builds connections with each encoder layer, so that the ABA-Net not only can exploit the \textit{coarse-grained} representations, of the source sequence, which is instrumental in language modeling, but also can exploit the \textit{fine-grained} representations of the source sequence, which is helpful to predict more precise answers. Hence, the answer predictor is encouraged to use source representations of different granularity, exploiting the expressive power of the model.
Furthermore, due to better textual feature representations are at the core for boosting MRC performance, we propose to
leverage the capsule network and self-attention module as the building blocks of encoders that separately encode the passage and question to obtain a more effective understanding of each word, i.e., the local and global representations. 
Last, before finally decoding to the probability that each position is the beginning or end of the answer span, we use our stacked encoder blocks to encode the generated representation again in the model encoding layer. Our experimental results show that each component has substantial gains in predicting answers correctly.

In general, our main contributions in this paper are as follows: 
\begin{itemize}

    \item We propose a novel approach called ABA-Net, which can encourage the answer predictor to take full advantage of the expressive power of the multi-layer encoder through exploring and exploiting the multi-granularity representations of source sequences.
    
    \item We propose to introduce the capsule network and self-attention module, which can better encode complicated text features  and can capture the global representations, as our basic encoder block for obtaining effective representations. Our encoder blocks also can be stacked to form a hierarchy for multi-step interactions, which can gradually capture mutual relation and better transfer knowledge.
    
     \item Specifically, the proposed approach achieves the state-of-the-art performance on the SQuAD 1.0 dataset, showing the concrete benefit of exploring and exploiting multi-granularity representations.
\end{itemize}

\section{Related Work}
\textbf{Machine Reading Comprehension.}\quad
In recent years, MRC \cite{DBLP:conf/icassp/ChenLYZZ21,chen2021good,DBLP:journals/corr/abs-2010-08923,DBLP:journals/corr/abs-2106-02182} has been an active field in NLP. Much of this popularity can be attributed to the release of many annotated and publicly available datasets, such as CNN/Daily News \cite{hermann2015teaching}, WikiReading \cite{hewlett2016wikireading}, SQuAD \cite{DBLP:conf/emnlp/RajpurkarZLL16}, TriviaQA \cite{joshi2017triviaqa},  DuReaderrobust  \cite{DBLP:journals/corr/abs-2004-11142}, Molweni \cite{DBLP:journals/corr/abs-2004-05080}, etc. 
Based on these challenging datasets, a great number of end-to-end approaches have been proposed, including BiDAF   \cite{DBLP:journals/corr/SeoKFH16}, DCN \cite{DBLP:journals/corr/XiongZS16}, R-Net  \cite{wang2017gated}. In MRC tasks, attention mechanism~\cite{DBLP:journals/access/DongLZWZ20,DBLP:journals/ijon/GaoWSL20,DBLP:journals/corr/abs-2005-05806,DBLP:journals/corr/abs-2001-09415,ijcai2021-549,chen-etal-2022-bridging} have become an essential part to capture dependencies without considering their distance in the input/output sequences. Recently, some works show that well pre-trained models are powerful and convenient for downstream tasks, such as R-Trans \cite{DBLP:journals/access/LiuZZW19}, DCMN+ \cite {DBLP:journals/corr/abs-1908-11511},  ALBERT \cite{DBLP:conf/iclr/LanCGGSS20} and GF-Net \cite{DBLP:journals/prl/LeeK20}, which facilitate us to take pre-trained models as our backbone encoder.

\textbf{Capsule Network.}\quad
Capsule Networks were initially proposed by Hinton~\shortcite{DBLP:conf/icann/HintonKW11}, which demonstrate the robustness in learning meaningful representations than vanilla CNNs. In \cite{DBLP:journals/corr/abs-1710-09829}, it refered to a capsule as a set of neurons whose activity vectors represent instantiation parameters of a particular type of entity, such as an object or part thereof.
Then Hinton~\shortcite{DBLP:conf/iclr/HintonSF18} introduced a novel iterative routing procedure between each capsule layer, which is based on the EM algorithm  and shows a higher level of accuracy  on the smallNORB  \cite{DBLP:conf/cvpr/LeCunHB04} dataset. Over the past few years, remarkable progress and research have been made with the capsule network in natural language processing (NLP). Dong~\shortcite{DBLP:journals/access/DongFWCDL20} 
employs Capsule Network based on BiLSTM for the sentiment analysis.
Zhao~\shortcite{DBLP:journals/corr/abs-1906-02829}, Wu~\shortcite{DBLP:journals/ijon/WuLWC20} and Yang~\shortcite{DBLP:journals/nn/YangZCQZS19} investigate Capsule Network for text classification.

Our work is different from our predecessors in the following aspects: (1) We propose a new attention mechanism to help MRC models leverage multi-granularity representations of each word reasonably ; (2) We innovatively combine the capsule network with self-attention in a residual encoder block to capture  the local and global representations, respectively ; (3) To the best of our knowledge, our model is the first to  apply the 
capsule network in MRC. 
\section{Model}

\subsection{Problem Formulation}
In this paper, the reading comprehension task is defined as follows: Given a passage/paragraph with $n$ words $P=\{p_1,p_2,p_3,...,p_n\}$ and a question with $m$ words  $Q=\{q_1,q_2,q_3,...,q_m\}$ as inputs, the goal is to find an answer span $A$ in $P$. If the question is answerable, the answer span $A$ exists in $P$ as a sequential text string; Otherwise,  $A$ is set to an empty string indicating that the question is unanswerable. Formally, the answer is formulated as $A=\{p_{begin} ,p_{end}\}$. In the case of unanswerable questions, $A$ denotes the last token of the passage.
\subsection{Model Overview}
Generally, our model contains four major parts: an embedding layer, an interaction-attention layer, a model encoding layer  and an output layer, as shown in Figure~\ref{fig:model}. $Embedding$ $layer$ is responsible for encoding each token in the passage and question into a fixed-length vector. In this layer, we use Glove \cite{DBLP:conf/emnlp/PenningtonSM14} as fundamental word embedding, and then they are passed through a BiLSTM \cite{hochreiter1997long} and our proposed Encoder block separately. Moreover, we utilize  ALBERT \cite{DBLP:conf/iclr/LanCGGSS20} contextualized word embedding.  $Interaction$-$Attention$ $layer$ plays a key role in ABA-Net, which  employs Adaptive Bidirectional Attention to distinguish and utilize multi-granularity  presentations of each word reasonably. $Model$ $Encoding$ $Layer$ encodes the generated representations again by our encoder blocks. $Output$ $Layer$ leverages the output of the previous layer to compute the final answer span by a bilinear projection.

\subsubsection{Embedding Layer}
\textbf{Word-Char Embedding.}
We obtain the fixed word embedding\footnote{We map the symbolic/surface feature of P and Q into neural space via word embeddings, 16-dim part-of-speech (POS) tagging embeddings, 8-dim named-entity embeddings and 4-dim hard-rule features.} from pre-trained Glove  \cite{DBLP:conf/emnlp/PenningtonSM14} word vectors.  Following \cite{DBLP:journals/corr/abs-1804-09541}, we obtain the character-level embedding of each word using CNN.  Following  \cite{DBLP:journals/corr/SeoKFH16}, a two-layer highway network in  \cite{DBLP:journals/corr/SrivastavaGS15} is used  on the top.  As results, we obtain the final embeddings for the tokens for P as a matrix $E_p\in \textbf{R}^{d\times{n}}$, and tokens in Q as a matrix $E_q\in \textbf{R}^{d\times{m}}$. 

 \textbf{Contextualized Word Embedding.}\quad Since the context encoding generated by the pre-trained model plays an important role in machine reading comprehension, we employ ALBERT \cite{DBLP:conf/iclr/LanCGGSS20} as contextualized embedding\footnote{Instead of adding a scoring layer to pre-trained models as proposed in many question answering models, we use the transformer output from ALBERT-large as contextualized embedding in our encoding layer.} 
like SDNET\cite{DBLP:journals/corr/abs-1812-03593}.  ALBERT generates L layers of hidden states for all BPE  \cite{DBLP:journals/corr/SennrichHB15} tokens in a sentence/passage and we employ a weighted sum of these hidden states to obtain contextualized embedding. In detail, given a word $w$, which is tokenized to $S$ BPE tokens $w={a_1,a_2,...,a_s}$, and ALBERT generates $L$ hidden states for each BPE token, $\bf{h_t^l}$, $1\leq{l}\leq{L}$, $1\leq{s}\leq{S}$. The contextual embedding ALBERT$_w$ for word $w$ is then a per-layer weighted sum of average ALBERT embedding, with weights $\theta_1,...,\theta_L$: 

\begin{equation}
 \textbf{ALBERT}_w= \sum_{l=1}^{L}\theta_{l} \frac{\sum_{s=1}^{S}\textbf{h}_t^l}{S}
\end{equation}

\textbf{BiLSTM.}\quad In  this  component, we  use  two  separate  bidirectional  RNNs  (BiLSTMs  \cite{hochreiter1997long}) to learn the contextualized understanding for $P$ and $Q$.
\begin{equation}
 	\textbf{h}_1^{P,k},...,\textbf{h}_m^{P,k}\!=\!BiLSTM(\textbf{h}_1^{P,k-1},... ,\textbf{h}_m^{P,k-1})
\end{equation}
\begin{equation}
 	\textbf{h}_1^{Q,k},...,\textbf{h}_m^{Q,k}\!=\!BiLSTM(\textbf{h}_1^{Q,k-1}\!,...,\textbf{h}_m^{Q,k-1})
\end{equation}
 where$1\leq{k}\leq{K}$ and $K$ is the number of BiLSTM layers. Thus, we get the representation:$\hat{C}_p\in\textbf{R}^{d\times{2n}}$ for passage and 
 $\hat{C}_p\in\textbf{R}^{d\times{2m}}$  for question.
 
\textbf{Encoder Block.}\quad In this component,
  we leverage and extend the strategy in  \cite{DBLP:journals/corr/abs-1804-09541} to construct our own encoder block architecture. As showed in Figure~\ref{fig:model}, we propose the following basic building block  : $[$convolution-PrimaryCaps-DigitCpas-layer $\times n$ + self-attention-layer + feed-forward-layer$]$ to encode the contextual information of both passages/paragraphs and questions  respectively.  In detail, the kernel size is 7, the number of filters is d = 128 and the number of Conv-Pri-Dig layers within a block is 5. Following \cite{DBLP:journals/corr/abs-1804-09541}, we adopt the multi-head attention mechanism defined in \cite{DBLP:journals/corr/VaswaniSPUJGKP17}, which calculates the weighted sum of all positions, for each position in the input, called a query or keys, based on the similarity between the query and key as measured by the dot product. Each of these basic operations (Conv-Pri-Dig/self-attention/ffn) is placed inside a {\em residual block} in  \cite{DBLP:journals/corr/abs-1804-09541}. For an input x and a given operation $f$, the output is $f(layernorm(x))+x$, meaning there is a full identity path from the input to output of each block.  As results, final representation is defined as: $C_p$ for passage and $C_q$ for questions.

 \begin{figure}[t]
\centering
\includegraphics[scale=0.5]{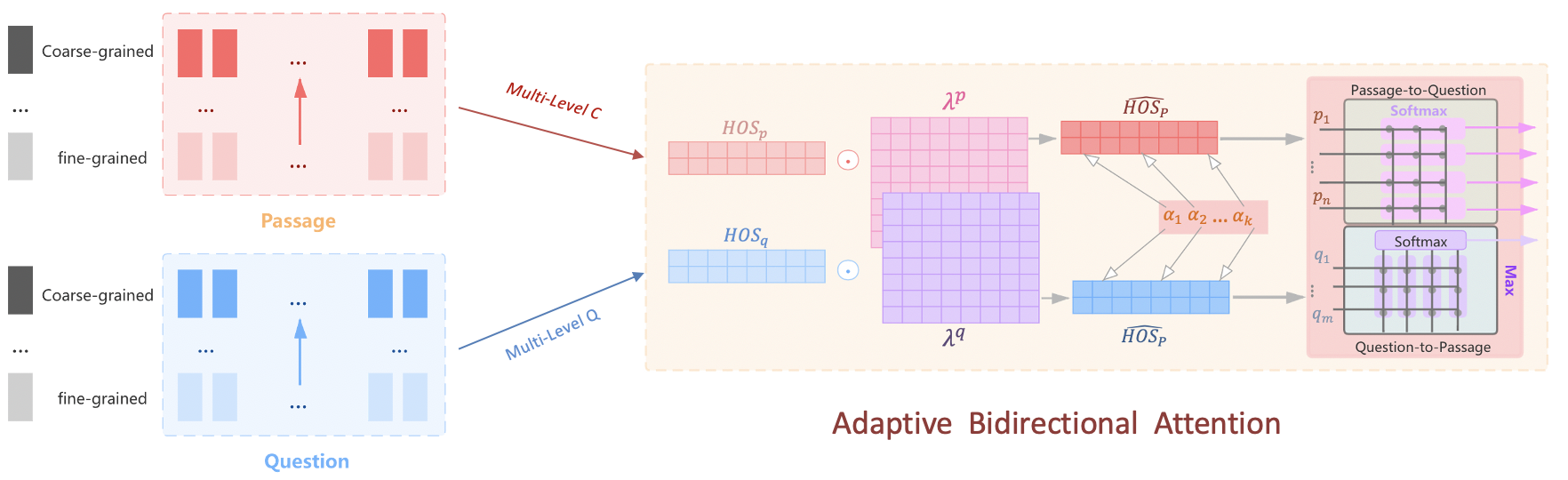}\\
\caption{An overview  of Adaptive Bidirectional Attention. $\odot$ is the element-wise multiplication.} \label{fig:attention} 
\end{figure}

 \subsubsection{Interaction-Attention Layer}
 \label{sec:interaction-attention}

\textbf{Adaptive Bidirectional Attention.}\quad After multiple layers  extract different levels of presentations  of each word, we conduct Adaptive Bidirectional attention from question to passage and passage to question based on all layers of generated representations to build connection of all layers of representation, which is showed in Figure~\ref{fig:attention}. We define a novel concept of 'history of semantic', which denotes multi-granularity representations extracted by the model before the current layer. Thus history-of-semantic vectors can be defined as:
 \begin{equation}
HOS_p=[C_w;C_c;E_p;\hat{ALBERT}_P;C_p;\hat{C}_p] \quad HOS_q=[Q_w;Q_c;E_q;\hat{ALBERT}_Q;C_q;\hat{C}_q]
\end{equation}
 where $C_w/Q_w$ and $C_c/Q_c$ are word/char embedding matrixs for passage and question respectively. 
 
 \textbf{Adaptive.}\quad In order to obtain multi-level representations   connection between the layers. We have designed the following function:
\begin{equation}
 \lambda ^{p} \times{HOS_p^{T}}=\widehat{HOS_p^{T}}  \quad \lambda ^{q} \times{HOS_q^{T}}=\widehat{HOS_q^{T}}	
 \end{equation}

where the matrix $\lambda$  is a trainable matrix. Notice that at the beginning of training, in order to retain the original semantics of each layer, we will initialize the first column of this matrix to all 1, the remaining columns are all 0.

However, since the data aggregation requires extensive computation resource, we specially train a set of selection parameters to reduce the dimension of  vectors: ${\alpha_1,\alpha_2,...,\alpha _{k}}$ to choose the three most relevant semantic vectors of $\widehat{HOS_p}$ and $\widehat{HOS_q}$. Thus, the input of bidirectional multilevel attention is defined as: $HOS^P$ for passages, $HOS^Q$ for questions.	

\textbf{Bidirectional  Attention.} \quad In this component,  our model  dynamically computes attention of the embeddings from previous layers each time , are allowed to flow through to the downstream modeling layer. Like most high performing models, such as  \cite{DBLP:journals/corr/XiongZS16}, \cite{DBLP:journals/corr/SeoKFH16},   \cite{DBLP:journals/corr/abs-1804-09541}, we construct passage-question and question-passage attention respectively. The attention function in  \cite{DBLP:journals/corr/SeoKFH16} is used to compute the similarity score between passages and questions. First, we calculate the similarities between each pair words in passages and questions, and rendering a similarity matrix $H\in \textbf{R}^{n\times{m}}$. H is computed as:
     \begin{equation}H=dropout(f_{attention}[HOS^P,HOS^Q])\end{equation}
  After that, we use the strategy of  \cite{DBLP:journals/corr/abs-1804-09541} to normalize each row of H by applying the softmax function, getting a matrix $ \hat H$. Then the passage-to-question attention is computed as $M= \hat H \cdot {HOS^Q}^T $.
  Following DCN \cite{DBLP:conf/iclr/XiongZS17} , we compute the column normalized matrix $\bar{H}$ of H by softmax function, and the question-to-passage attention is $S = \hat H \cdot \bar H^T \cdot {HOS^P}^T$. At last, our model use a simple concatenation as following to get the final representation, which shows good performance in our experiments: $O=[{HOS^P};M;{HOS^P} \odot M;{HOS^P} \odot S]$.
 \subsubsection{Model Encoding Layer}\quad The input of this layer is O, which encodes the question-aware representations of context words. This layer use the same building blocks as the Interaction Layer except that Conv-Pri-Dig layer number is 2 within a block and the total number of blocks are 4. Following \cite{DBLP:journals/corr/abs-1804-09541},  we share weights between each of the 3 repetitions of the model encoder.
 
 \subsubsection{Output Layer}\quad This layer aims to calculate two scores for each word in the passage, corresponding to the probability that the answer begins and ends at that word. Like most existing models, the last output layer is task-specific. We adopt the following strategy \cite{DBLP:journals/corr/abs-1804-09541} to predict the probability of each position in the passage/paragraph being the start or end of an answer span, if the question is answerable. More concretely, the distributed probabilities of the starting and ending index for the answer span are defined as:
\begin{equation}
P^{begin}=softmax(W_1[B_1;B_2])
\end{equation}
\begin{equation}
P^{end}=softmax(W_2[B_2;B_3])
\end{equation}
where $W_1$ and $W_2$ are two trainable variables and $B_1$, $B_2$, $B_3$ are respectively the outputs of the three model encoders, from bottom to top. If the question is unanswerable, the final output is an empty string. In addition, the proposed model can be customized to other comprehension tasks, e.g. selecting from the candidate answers, by changing the output layers accordingly.
  
\textbf{Objective }\quad Following \cite{DBLP:journals/corr/abs-1804-09541}, the objective function is defined as the negative sum of the log probabilities of the predicted distributions indexed by true start and end indices, averaged over all the training examples:
 \begin{equation}	L(\theta)=-\frac{1}{N}\sum_{i}^{N}[log(p^{begin}_{y^1_i})+log(p^{end}_{y^2_i})]
\end{equation}
  where $y_i^1$and $y_i ^2$are respectively the ground truth starting and ending position of example i, and contains all the trainable variables.
%
%
    %
    %
    %
    %
    %

%
%

\section{Experiments }

In this section, we conduct experiments to study the performance of our model. We will primarily benchmark our model on the SQuAD 2.0 dataset \cite{DBLP:conf/acl/RondeauH18} and SQuAD 1.0 dataset  \cite{DBLP:conf/emnlp/RajpurkarZLL16}. We also conduct similar studies on COQA dataset \cite{reddy-etal-2019-coqa}, a conventional Q\&A dataset, to show that the effectiveness and efficiency of our model are general. Detailed introduction of 
these datasets and experimental settings can be found in Appendix A.

\begin{table}
\small
\renewcommand{\arraystretch}{1.3}
    \centering
    \begin{tabular}{l c c c}
    \hline
    \hline
    & SQuAD 1.0 &SQUAD 2.0 & COQA \\
    \hline
    Single Model & EM/F1 & EM/F1 & F1(average)\\
    \hline


        SAN \cite{liu2018stochastic} & 76.8/84.2 & 68.6/71.4&-\\
   BiDAF++ \cite{DBLP:conf/iclr/ShenZL0Z18} & 77.6/84.9 &65.6/68.7& 69.5\\
    QANet \cite{DBLP:journals/corr/abs-1804-09541} & 80.9/87.8 & 65.4/67.2&-\\
    
   Bert$_{large}$ \cite{DBLP:journals/corr/abs-1810-04805} & 85.1/91.8 & 79.9/83.1 & 74.4\\
   SDNet \cite{DBLP:journals/corr/abs-1812-03593} &- &76.7/79.8& 76.6\\
   SGNet \cite{DBLP:conf/aaai/0001WZDZ020} &-& 85.1/87.9&-\\
   SemBERT$_{large}$ \cite{zhang2019semantics} &- & 86.1/88.8&-\\
   RoBERTa $_{large}$\cite{liu2019roberta} &- & 86.8/89.8&84.9\\
  
   ALBERT$_{large}$ \cite{lan2019albert} & 89.1/94.6 & 86.8/89.6& 85.4\\
   XLNet$_{large}$ \cite{yang2019xlnet} & 89.9/95.0 & 87.9/90.7 & 84.6\\
   Retr-Reader on ELECTRA \cite{zhang2020retrospective}&-& 89.5/\textbf{92.0}&-\\
   TR-MT ($WeChatAI$,2019) & -&-& \bfseries{89.3}\\
    \hline
      Dev :ABA-Net& \bfseries{90.1/95.4} & 88.8/91.3& 88.4 \\
      \cdashline{1-4}
       Test :ABA-Net&\bfseries{90.4/95.8} & \textbf{89.6} {/91.7} & 88.5\\
       \hline
       \hline
    \\
    \end{tabular}
\caption{The performances of single models on different  datasets. }
\label{tab:my_label_1}
\end{table}
\subsection{Results }
The F1 and Exact Match (EM) are two metrics of accuracy for evaluating MRC models performance. F1 measures the part of the overlapping mark between the predicted answer and groundtruth, and if the prediction is exactly the same as the ground truth, the exact match score is 1, otherwise it is 0. We show the results in comparison with other single models in Table~\ref{tab:my_label_1}. From the table, the accuracy (EM/F1) performance of our model is on par with the state-of-the-art models.  In detail, ABA-Net achieves the state-of-the-art results on SQuAD 1.0 and competitive performance on SQuAD 2.0 and COQA compared with other baseline models. More concretely, ABA-Net improves EM/F1 score to 90.4/95.8, compared with previous state-of-the-art models on SQuAD 1.0; ABA-Net also achieves an EM score of 89.6 and an F1 score of 91.7 on SQuAD 2.0. Furthermore, our model achieves an F1 (average) score of 88.5 on COQA, which is competitive with the best documented result.
\begin{table}[]
\begin{center}
\resizebox{\textwidth}{!}{
    \begin{tabular}{l c  c |c c |c c}
    \hline
     \hline
     & SQuAD 1.0 & Diff to B-M & SQuAD 2.0 & Diff to B-M  & COQA & Diff to B-M\\    
         &EM/F1 &EM/F1  &EM/F1 &EM/F1  &EM/F1 &F1(average)\\
    \hline
    Base ABA-Net & \textbf{90.2/95.4} & & \textbf{89.6/91.7}&&\textbf{88.4} \\
    \hline
    - self-attention in encoders & 88.8/94.2 & \bfseries{-1.4}/\bfseries{-1.2} &
    88.0/90.4 &\bfseries{-1.6}/\bfseries{-1.3} & 87.0 &\bfseries{-1.4}    \\
- Capsule Network in encoders & 88.7/93.7 & \bfseries{-1.5}/\bfseries{-1.7} & 88.5/90.4 & \bfseries{-1.1}/\bfseries{-1.3} & 87.3 & \bfseries{-1.1} \\
- \bfseries{Adaptive Bidirectional Attention} & 88.6/93.3 & \bfseries{-1.6}/\bfseries{-2.1}
 & 87.8/89.5& \bfseries{-1.8}/\bfseries{-2.2} & 85.9 & \bfseries{-2.5} \\
replace $sep^{\clubsuit}$ $conv$ with normal conv & 89.7/95.0 & \bfseries{-0.5}/\bfseries{-0.5}
& 89.2/91.2 & \bfseries{-0.4}/\bfseries{-0.5}& 87.9 & \bfseries{-0.3} \\
    \hline
    \hline
    \\
    \end{tabular} }
    \end{center}
    \caption{An ablation study of our model. The reported results are obtained on the $test$  set and we use the original ABA-Net as the base model. $Diff$ $to$ $B$-$M$ is the abbreviation of $Difference$ $ to $ $Base$ $Model$.  $sep^{\clubsuit}$ $conv$ is the abbreviation of $depthwise$ $separable$ $convolutions$. }   
    \label{tab:my_label_2}
    \end{table}

\subsection{Ablation Study And Analysis}
\raggedbottom
\textbf{Ablation Study.} \quad We conduct ablation studies on components of the proposed model. The validation scores on the test set are shown in Table~\ref{tab:my_label_2}. As can be seen from the table, the Adaptive Bidirectional Attention plays a crucial part in our model: both F1 and EM drop dramatically by almost 2 percent if it is removed. Self-attention in the encoders is also a necessary component that contributes 1.4/1.2,1.6/1.3,1.4 to the ultimate performance on three datasets. More significantly, the combination of Capsule Network  and self-attention is significantly better than the combination of RNNs or CNNs and self-attention. 

\textbf{Analysis. }\quad We interpret these phenomena as follows:  Adaptive Bidirectional Attention is able to distinguish and utilize multi-granularity  representations of each word reasonably. The capsule network captures the local structure of the context and questions while the self-attention is able to model the global context representations, interactions  and further focus on the most important part of the interaction. Hence they are complimentary to but cannot replace each other. Concretely, capsule  networks have the ability for learning hierarchical relationships between consecutive layers by using routing processes without parameters and additionally improve the generalization capability. The use of separable convolutions instead of traditional convolutions also has a prominent contribution to performance, which can be seen by replacing the depthwise separable convolutions with conventional convolutions, resulting in slightly worse accuracy.
\section{Discussions \& Visualization }
\begin{figure}
\makeatletter\def\@captype{figure}\makeatother
\begin{minipage}{0.47\textwidth}
\centering
\includegraphics[scale=0.45]{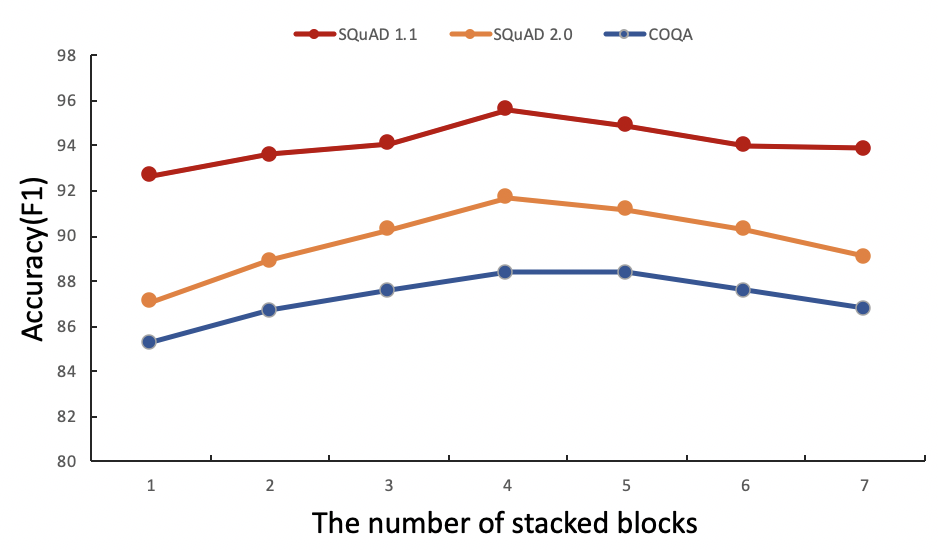}
\caption{
F1 accuracy of various stacked encoder blocks on three benchmark datasets.}\label{fig:zhe} 
\end{minipage}
\hfil
\makeatletter\def\@captype{table}\makeatother
\begin{minipage}{0.48\textwidth}
\centering
        \footnotesize
        \begin{tabular}{l c}
         \hline
    \hline
         & EM/F1 \\
    \hline
    Base Model & \\
    \hline
    BiDAF++ \cite{DBLP:conf/iclr/ShenZL0Z18}& 65.6/68.7 \\
    QANet \cite{DBLP:journals/corr/abs-1804-09541}& 65.4/67.2 \\
    SAN \cite{liu2018stochastic} & 68.6/71.4 \\  
    SDNet \cite{DBLP:journals/corr/abs-1812-03593}& 76.7/79.8 \\
    SGNet\cite{DBLP:conf/aaai/0001WZDZ020} & 85.1/87.9 \\
  \hline
 + Adaptive Bidirectional Attention \\
  \hline
     BiDAF++ \cite{DBLP:conf/iclr/ShenZL0Z18} & \bfseries{67.2/70.0} \\
    QANet \cite{DBLP:journals/corr/abs-1804-09541}& \bfseries {66.9/69.1} \\
    SAN \cite{liu2018stochastic}& \bfseries {70.1/73.0} \\
    SDNet\cite{DBLP:journals/corr/abs-1812-03593} & \bfseries{78.5/81.1} \\
    SGNet \cite{DBLP:conf/aaai/0001WZDZ020} & \bfseries{86.7/89.2} \\
    \hline
    \hline \\
    \end{tabular}
    \vspace{-2\baselineskip}

    \caption{An comparison study of Adaptive Bidirectional Attention on SQuAD 2.0. }
    \label{tab:my_label_3}
    \end{minipage} 
\end{figure}


\textbf{Effect of Adaptive Bidirectional Attention .}\quad We additionally perform experiments to prove the effectiveness of our proposed Adaptive Bidirectional Attention mechanism. Specifically, we add Adaptive Bidirectional Attention on some end-to-end MRC models to compare with their initial version (i.e. base models in Table~\ref{tab:my_label_3}) on SQuAD 2.0. As can be seen from Table~\ref{tab:my_label_3}, after adding Adaptive Bidirectional Attention, the performance of these models could be improved to varying degrees. This also proves the versatility of this attention mechanism. 

\textbf{Effect of Capsule Network .}\quad We additionally perform experiments to understand this mechanism, which is shown in Table ~\ref{tab:my_label_4}. We first adopt normal convolutions to replace the Capsule Network and the effect of the model decreases heavily on three benchmark  datasets. We attribute it to the fact that Capsule Network is better than CNNs in extracting rich text representation. Then we employ a shared two-layers BiLSTM to replace the Capsule Network, which gives 0.9/1.1 and 0.6/1.0 EM/F1 decrease, 0.3 F1 decrease in our experiments.  Notice that, due to the limitations of the existing 
capsule network, its performance on the complex multi-round dialogue reading comprehension dataset(COQA) does not perform better than other two  datasets.

\textbf{Effect of the total number of stacked blocks .}\quad Our model uses stacked encoder blocks to form a multi-step interactions. As showed in Figure~\ref{fig:zhe}, we observe that the performance of our model drops heavily when encountered with different number of blocks. Under the premise that the number of blocks is 4, ABA-Net achieves the best results on three datasets. We interpret this phenomena as follows: When the number of encoder blocks is less than 4, the local and global text features captured by the model are not enough, and it is easy to overfit. On the contrary, if the number exceeds, the model extracts many repeated contextual 
representations , which will hurt the performance.

\begin{figure}[t]
\centering
\includegraphics[scale=0.24]{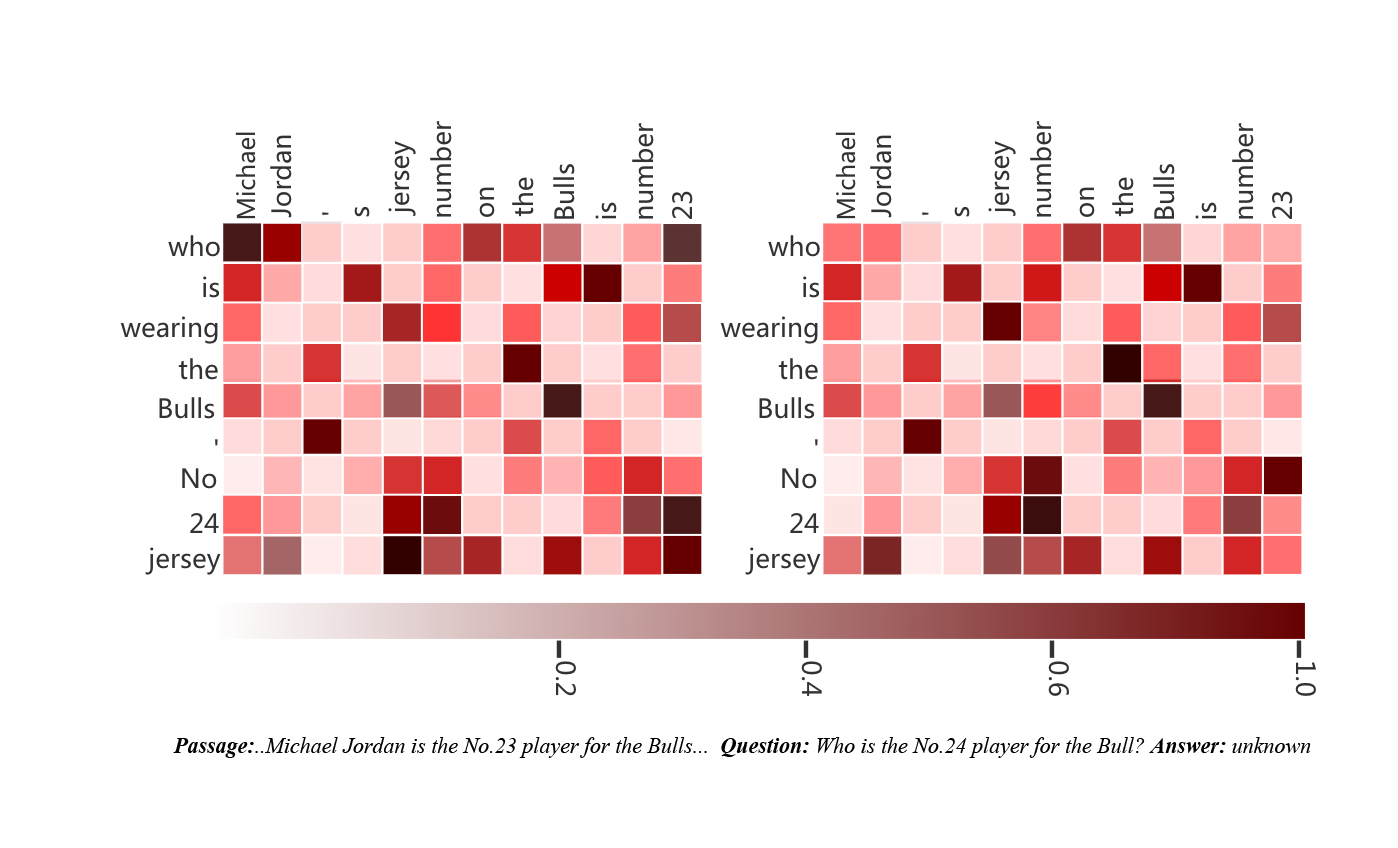}\\
\caption{
Visualization of Syntax-guided attention (left) and Adaptive Bidirectional attention (right). For Syntax-guided attention, $who$ focuses on $Michael$ and \textit{\textbf{23}}, and there is a great similarity between \textit{\textbf{23}} and \textit{\textbf{24}}, which would mislead the model to make wrong judgment. In contrast, this phenomenon does not  happen in our Adaptive Bidirectional attention. } \label{fig:attention1}
\end{figure}

\textbf{ Visualization.}\quad To have an insight that how Adaptive Bidirectional Attention works, we draw attention distributions of the Syntax-guided attention of SGNet \cite{DBLP:conf/aaai/0001WZDZ020} and our proposed Adaptive Bidirectional attention, as shown in Figure~\ref{fig:attention1}. The visualization verifies that benefiting from Adaptive Bidirectional attention , our model is effective at  distinguishing and utilizing different layers of presentation of each word in the context and the query , guiding the downstream layer to collect more relevant semantics to make predictions.
\begin{table}[]
    \centering
    \footnotesize
    \begin{tabular}{l c c c}
    \hline
    \hline
        & SQuAD 1.0 & SQuAD 2.0 & COQA\\
        \hline
         The Effect of Capsule Network &EM/F1&EM/F1&F1\\
    \hline
    Base ABA-Net & \textbf{90.2/95.6} & \textbf{89.6/91.7} & \textbf{88.5}  \\
    \hline
    replace Capsule with normal conv & 88.3/93.6 & 88.4/90.3 & 87.5\\
   replace Capsule with sep conv & 88.8/94.1 &  88.8/90.8 & 88.0\\
    replace Capsule with BiLSTM &89.3/94.5 & 89.0/90.7 & 88.1 \\
    \hline
    \hline
    \\
    \end{tabular}
    \caption{An comparison   study of the Capsule Network . }  
    \label{tab:my_label_4}
\end{table}

\section{Conclusion}
In this paper, we propose a novel end-to-end machine reading comprehension (MRC) model Adaptive Bidirectional Attention-Capsule Net-work (ABA-Net) for exploring and exploiting multi-granularity representations of source sequences. In particular, our ABA-Net can adaptively exploit the source representations of different levels to the predictor.  More concretely, our core innovation is to propose a novel attention mechanism called Adaptive Bidirectional Attention, which guides attention learning and  reasonably leverages source representations of different levels for question answering by dynamically capturing the connection between all layers of representations of each word.  Furthermore, to explore better representations, we incorporate the capsule network and self-attention module as the building blocks of encoders. The experimental results show that our model achieves competitive performance compared to the state-of-the-art on three public benchmark datasets. The systematic analysis demonstrates the effectiveness of each component in our model and shows that our model can learn effective representations for MRC and is capable of understanding rich context and answering complex questions. Future work can include an extension of employing Adaptive Bidirectional Attention to other question answering tasks.

\bibliographystyle{coling}
\bibliography{coling2020}

\end{document}


\maketitle

\appendix
\section{Dataset And Experimental Settings}
\subsection{Dataset}  SQuAD 1.0 \cite{DBLP:conf/emnlp/RajpurkarZLL16} contains 107,785 questions and 536 accompanying articles. The typical length of the paragraphs is around 250 while length of the question is 10 masks, although there are exceptions to this.    
We also evaluate our model on SQuAD 2.0 dataset \cite{DBLP:conf/acl/RondeauH18}, a new MRC dataset which is a combination of SQuAD 1.0 and additional unanswerable question-answer pairs. Specifically, the number of questions that can be answered is about 100K; while the unanswerable questions are around 53K. For the SQuAD 2.0 challenge, MRC models must answer questions not only where possible, but also when the paragraph does not support any answers. As a widely benchmark conversational question answering dataset, CoQA \cite{reddy-etal-2019-coqa} requires MRC systems to understand dialogue context, which consists of more than 127k questions with answers collected from 8k conversations.  Two evaluation metrics are used: Exact Match (EM) and Macro-averaged F1 score (F1) \cite{DBLP:conf/emnlp/RajpurkarZLL16}.
 
\subsection{Implementation Details} Similar to \cite{DBLP:journals/corr/abs-1804-09541}, the strategies of standard regularizations are as follows: First, we use dropout on character, word embeddings and between layers, where the dropout rates of the word and character  are 0.1 and 0.05 respectively, and the dropout rate between every two layers is 0.1. In addition, we use L2 weight decay on all the trainable matrices. We also adopt the stochastic depth method (layer dropout) \cite{DBLP:conf/kdd/ShenHGC17} within each embedding or model encoder layer, where sublayer l has survival probability $p_l=1-\frac{l}{L}(1-p_L)$  where L is the last layer and $p_L= 0.9$.

The word embeddings used  for  questions  and passages  are  initialized from 300-dimensional  GloVe  vectors and fixed during training.\footnote{All the out-of-vocabulary words are mapped to an $<$\texttt{UNK}$>$ token.} 
The hidden size and the convolution filter number of encoder block in the interaction layer are all 128, and the batch size is 25. Regarding the hidden size of our model, we search greedily among \{128, 256, 300\}. The numbers of Conv-Pri-Dig layers in the embedding and modeling encoder are 5 and 2, kernel sizes are 7 and 5, and the block numbers for the encoders are 1 and 4, respectively.

\bibliographystyle{coling}
\bibliography{coling2020}